\documentclass[journal,onecolumn]{IEEEtran}
\usepackage[english]{babel}
\usepackage[utf8]{inputenc}
\usepackage{amsmath}
\usepackage{graphicx}
\usepackage{xparse}
\usepackage{setspace}
\usepackage{array}
\usepackage[dvipsnames]{xcolor}
\usepackage{amssymb}

\usepackage{enumitem}
\newcommand{\subscript}[2]{$#1  #2$}

\DeclareMathOperator*{\argmin}{arg\,min}

\newtheorem{Theorem}{\textbf{Theorem}}
\newtheorem{corollary}{\textbf{Corollary}}

\begin{document}

\title{\huge Optimizing Pipelined Computation and Communication for Latency-Constrained Edge Learning}

\author{Nicolas~Skatchkovsky,~\IEEEmembership{Student Member,~IEEE,}
        and~Osvaldo~Simeone,~\IEEEmembership{Fellow,~IEEE}
\thanks{The authors are with the Department of Informatics, King's College London, United Kingdom (e-mails: \{nicolas.skatchkovsky, osvaldo.simeone\}@kcl.ac.uk).}
\thanks{The authors have received funding from the European Research Council (ERC) under the European Union’s Horizon 2020 Research and Innovation Programme (Grant Agreement No. 725731).}
}

\maketitle

\begin{abstract}
Consider a device that is connected to an edge processor via a communication channel. The device holds local data that is to be offloaded to the edge processor so as to train a machine learning model, e.g., for regression or classification. Transmission of the data to the learning processor, as well as training based on Stochastic Gradient Descent (SGD), must be both completed within a time limit. Assuming that communication and computation can be pipelined, this letter investigates the optimal choice for the packet payload size, given the overhead of each data packet transmission and the ratio between the computation and the communication rates. This amounts to a tradeoff between bias and variance, since communicating the entire data set first reduces the bias of the training process but it may not leave sufficient time for learning. Analytical bounds on the expected optimality gap are derived so as to enable an effective optimization, which is validated in numerical results.
\end{abstract}


\section{Introduction}

\begin{figure}
    \centering
    \includegraphics[scale=0.18]{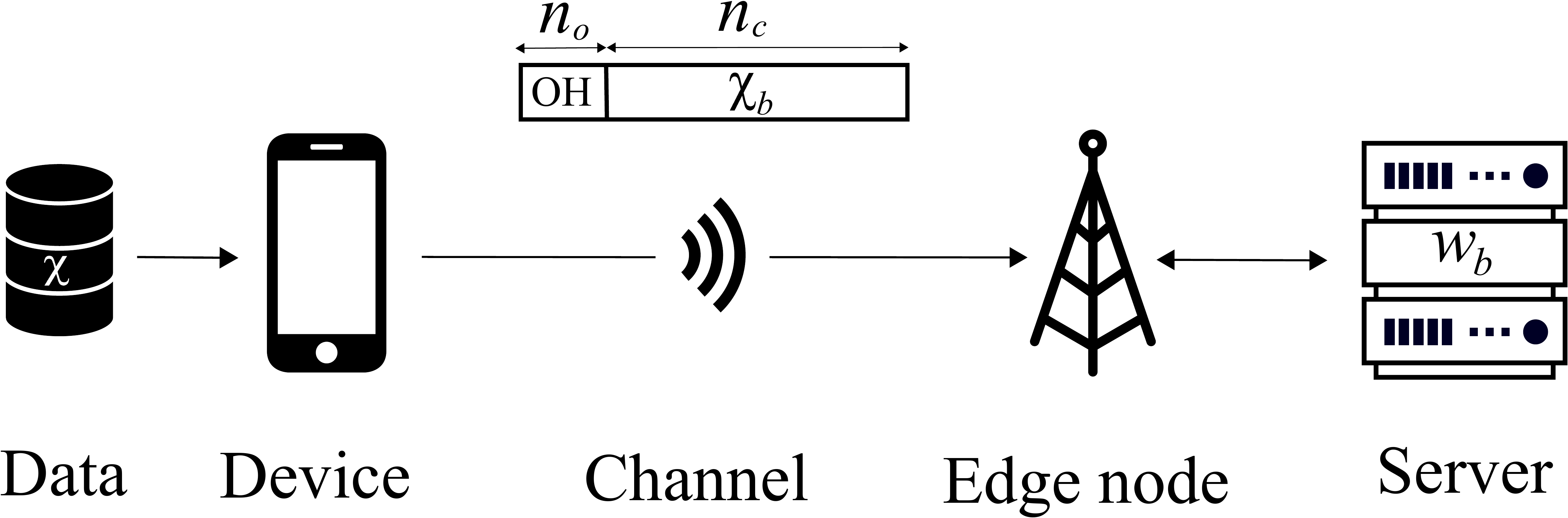}
    \caption{An edge computing system, in which training of a model parametrized by vector $w$ takes place at an edge processor based on data received from a device using a protocol with timeline illustrated in Fig. \ref{fig 2} (OH = overhead).}
    \label{fig 1}
\end{figure}

\IEEEPARstart{E}{dge} learning refers to the training of machine learning models on devices that are close to the end users \cite{DBLP:journals/corr/abs-1812-02858}. The proximity to the user is instrumental in facilitating a low-latency response, in enhancing privacy, and in reducing backhaul congestion. Edge learning processors include smart phones and other user-owned devices, as well as edge nodes of a wireless network that provide wireless access and computational resources \cite{DBLP:journals/corr/abs-1812-02858}. As illustrated in Fig. \ref{fig 1}, the latter case hinges on the offloading of data from the data-bearing device to the edge processor, and can be seen as an instance of mobile edge computing \cite{ETSI:MECWhitePaper}.

Research on edge learning has so far instead focused mostly on scenarios in which training occurs locally at the data-bearing devices. In these setups, devices can communicate either through a parameter server \cite{DBLP:journals/corr/abs-1811-03748} or in a device-to-device manner \cite{Wang}. The goal is to either learn a global model without exchanging directly the local data \cite{Teerapittayanon} or to train separate models while leveraging the correlation among the local data sets \cite{NIPS2017_7029}. Devices can exchange either information about the local model parameters, as in federated learning \cite{DBLP:journals/corr/McMahanMRA16}, or gradient information, as in distributed Stochastic Gradient Descent (SGD) methods \cite{DBLP:journals/corr/abs-1901-00844, bottou}.

In this work, we consider an edge learning scenario in which training takes place at an edge node of a wireless system as illustrated in Fig. 1. The data is held by a device and has to be offloaded through a communication channel to the edge node. The learning task has to be executed within a time limit, which might be insufficient to transmit the complete dataset. Transmission of data blocks from device to edge node, and training at the edge node can be carried out simultaneously (see Fig. \ref{fig 2}). Each transmitted packet contains a fixed overhead, accounting e.g. for meta-data and pilots. Given the overhead of each data packet transmission, what is the optimal size of a communication block?  Communicating the entire data set first reduces the bias of the training process but it may not leave sufficient time for learning. We investigate a more general strategy that communicates in blocks and pipelines communication and computation with an optimized block size, which is shown to be generally preferable. Analysis and simulation results provide insights into the optimal duration of the communication block and on the performance gains attainable with an optimized communication and computation policy. 

The rest of this letter is organized as follows. In Sec. \ref{sec2}, we provide an overview of the model and the associated notations. In Sec. \ref{sec3}, we examine the technical assumptions necessary for our work. In Sec. \ref{sec4}, we provide our main result and discuss its implications. Finally, in Sec. \ref{simus}, we consider numerical experiments in the light of our result. 

\section{System model}
\label{sec2}

\begin{figure*}
    \centering
    \includegraphics[scale=0.6]{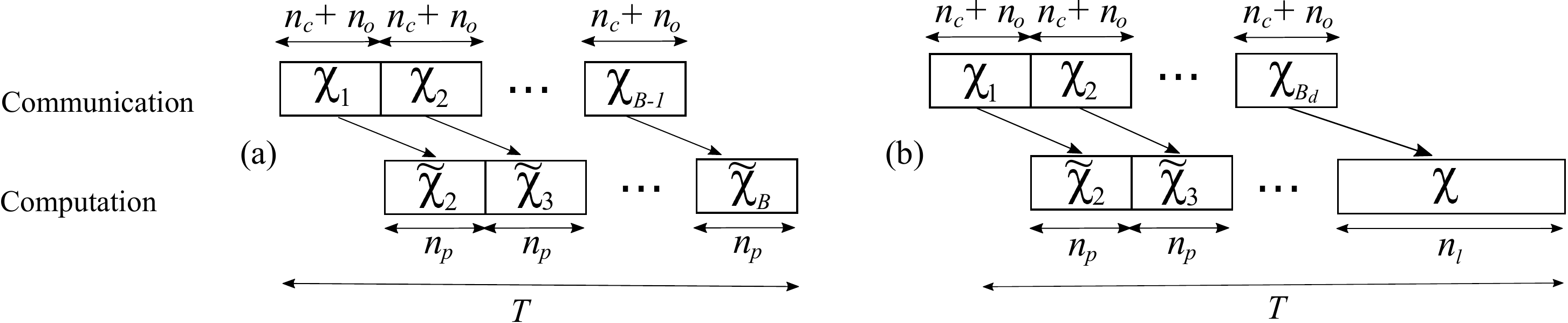}
    \caption{Transmission and training protocol: when (a) $T \leq B_{d}(n_{c} + n_{o})$; and (b) $T > B_{d}(n_{c} + n_{o})$.}
    \label{fig 2}
\end{figure*}

As seen in Fig. \ref{fig 1}, we study an edge learning system in which a device communicates with an edge node, and associated server, over an error-free communication channel. The device has access to a local training dataset $\mathcal{X} = \{x_{1}, x_{2}, \dots, x_{N}\}$ of $N$ data points $\{x_{n}\}_{n=1}^{N}$, and training of a machine learning model is carried out at the edge node based on data received from the device. As illustrated in Fig. \ref{fig 2}, communication and learning must be completed within a time limit $T$. To this end, the transmissions are organized into blocks, and transmission and computing at the edge node can be performed in parallel. 

Training at the edge node aims at identifying a model parametrized by a vector \small $w \in \mathrm{R}^{d}$ \normalsize within a given hypothesis class. Training is carried out by (approximately) solving the Empirical Risk Minimization (ERM) problem (see, e.g, \cite{simeone}). This amounts to the minimization with respect to vector $w$ of the empirical average \small $\mathcal{L}(w)$ \normalsize of a loss function $\ell(w,x)$ over all the data points $x$ in the training dataset, i.e.,

\small
\begin{equation}
\label{ERM}
    \mathcal{L}(w) = \frac{1}{N} \sum_{n=1}^{N}\ell(w, x_{n}).
\end{equation}
\normalsize
As detailed below, the minimization of the function $\mathcal{L}(w)$ is carried out at the edge node using SGD, based on the data points received from the device.

In order to elaborate on the communication and computation protocol illustrated in Fig. \ref{fig 2}, we normalize all time measures to the time required to transmit one data sample from the device to the edge node. With this convention, we denote as $\tau_{p}$ the time required to make one SGD update at the edge node. 

As seen in Fig. \ref{fig 2}, transmission from the device to the edge node is organised into blocks. In this study, we ignore the effect of channel errors, which is briefly discussed in Sec. \ref{sec6}. In the $b$-th block, the device transmits a subset $\mathcal{X}_{b} \subseteq \mathcal{X}$ of $n_{c}$ new samples from its local dataset. At the end of the block, the edge node adds these samples to the subset \small $\tilde{\mathcal{X}}_{b+1}$ \normalsize of samples it has available for training in the $b+1$-th block, i.e., \small $\tilde{\mathcal{X}}_{b+1} = \tilde{\mathcal{X}}_{b} \cup \mathcal{X}_{b}$ \normalsize with \small $\mathcal{X}_{0} = \varnothing.$ \normalsize The samples in \small $\mathcal{X}_{b}$ \normalsize are randomly and uniformly selected from the set \small $\Delta \mathcal{X}_{b} = \mathcal{X} \setminus \tilde{\mathcal{X}}_{b}$ \normalsize of samples not yet transmitted to the edge node. 
A packet sent in any block contains an overhead, e.g., for pilots and meta-data, of duration $n_{o}$, irrespective of the number $n_{c}$ of transmitted samples. It follows that the duration of a transmission block is $n_{c} + n_{o}$. 

There are at most $B_{d} = N/n_{c}$ transmission blocks, since $B_{d}$ blocks are sufficient to deliver the entire dataset to the edge node. Therefore, we need to distinguish two cases. As seen in Fig. \ref{fig 2}(a), when $T \leq B_{d}(n_{c} + n_{o})$, the device is only able to deliver a fraction of the samples. In particular, denoting as $B= T/(n_{c} + n_{o})$ the number of blocks, the fraction of data points delivered at the edge node at time $T$ equals $(B - 1)/B_{d}$. In contrast, if $T > B_{d}(n_{c} + n_{o})$, as illustrated in Fig. \ref{fig 2}(b), the edge node has the entire dataset available after $B_{d}$ blocks, that is, for a duration equal to $ \tau_{l} = T - B_{d}(n_{c} + n_{o})$. Henceforth, we refer to this last period as block  $B_{l} = B_{d} + 1$.

During each block $b \leq B_{d}$, the edge node computes $n_{p} = (n_{c} + n_{o})/\tau_{p}$ local SGD updates (\ref{SGD}). During block $B_{l}$, the edge node computes $n_{l} = \tau_{l}/\tau_{p}$ SGD updates. The $j$-th local update at block $b$, with $j = 1,\dots, n_{p}$, is given as
\small
\begin{equation}
    \label{SGD}
    w_{b}^{j} = w_{b}^{j-1} - \alpha \nabla \ell(w_{b}^{j-1}, \xi_{b}^{j}),
\end{equation}
\normalsize
where $\alpha$ is the learning rate, and $\xi_{b}^{j}$ is a data point sampled i.i.d. uniformly from the subset $\tilde{\mathcal{X}}_{b} = \bigcup_{l=1}^{b-1} \mathcal{X}_{l}$ of samples currently available at the edge node. Note that we have $\tilde{\mathcal{X}}_{B_{l}} = \mathcal{X}$.

The goal of this work is  to optimize the number of samples $n_{c}$ sent in each block with the aim of minimizing the empirical loss (\ref{ERM}) at the edge node at the end of time $T$. In the next sections, we present an analysis of the empirical loss obtained at time $T$ that allows us to gain insights into the optimal choice of $n_{c}$.

\section{Technical assumptions}
\label{sec3}
In order to study the training loss achieved at the edge node at the end of the training process, we make the following standard assumptions, which apply, for instance, to linear models with quadratic or cross-entropy losses under suitable constraints (see the comprehensive review paper \cite{bottou}):
\begin{enumerate}[label=(\subscript{A}{\arabic*}), align=left, leftmargin=0.5\parindent]
\setlength{\itemsep}{5pt}
    \item \label{ass1} the sequence of iterates $w_{b}^{j}$ in (\ref{SGD}) is contained in a bounded open set $\mathcal{W} \subseteq \mathbf{R}^{d}$ with radius $D = \max_{u, w \in \mathcal{W} \times \mathcal{W}} ||w - u||_{2}$ over which the function $\ell(w, x)$ is bounded below by a scalar $\ell_{\text{inf}}$ for all $x$;
    
    \item \label{ass2}  the  function  $\ell(w,x)$ is continuously differentiable in $w$ for any fixed value of $x$ and is $L$-smooth in $w$, i.e.,
    \small
    \begin{align}
    ||\nabla \ell(w, x) - \nabla \ell(\Bar{w}, x)||_{2} \leq L||w - \Bar{w}||_{2}
    \end{align}
    \normalsize
    for all $(w, \Bar{w}) \in \mathcal{W}\times\mathcal{W}$, and for all $x$. This implies
    \small
    \begin{align}
    \ell(w, x) \leq \ell(\Bar{w}, x) + \nabla \ell(\Bar{w}, x)^{T}(w - \Bar{w}) + \frac{L}{2}||w - \Bar{w}||_{2}^{2}
    \end{align}
    \normalsize
    for all $(w, \Bar{w}) \in \mathcal{W} \times \mathcal{W}$, and for all $x$;
    \item \label{ass3} the loss function $\ell(w,x)$ is convex and satisties the Polyak-Lojasiewicz condition in $w$, i.e., there exists a constant $c > 0$ such that
    \small
    \begin{align}
    2c(\ell(w, x) - \ell(w_{\ell}^{*}, x)) \leq ||\nabla \ell(w, x)||_{2}^{2} 
    \end{align}
    \normalsize
    for all $(w, x) \in \mathcal{W} \times \mathbf{R}^{d}$ where $w_{\ell}^{*}(x) = \argmin_{w \in \mathcal{W}}\ell(w,x)$ is a minimizer of $\ell(w, x)$. The P-L condition is implied by, but does not imply, strong convexity \cite{bottou}.
    \end{enumerate}

We further need to make assumptions on the statistics of the gradient $\nabla \ell(w, \xi_{b}^{j})$ used in the update (\ref{SGD}). To this end, for each block $b > 1$, we define the empirical loss limited to the samples available at the edge node at block $b$ as
\small
\begin{equation}
\label{loss1}
\tilde{\mathcal{L}}_{b}(w) = \frac{1}{(b - 1) n_{c}} \sum \limits_{x_{i} \in \tilde{\mathcal{X}}_{b}}\ell(w, x_{i});
\end{equation}
\normalsize
the empirical loss over the samples transmitted at iteration $b \geq 1$ as
\small
\begin{equation}
\label{loss2}
\mathcal{L}_{b}(w) = \frac{1}{n_{c}}\sum \limits_{x_{i} \in \mathcal{X}_{b}}\ell(w, x_{i});
\end{equation}
\normalsize
and the empirical loss over the samples not available at the edge at iteration $b > 1$
\small
\begin{equation}
\label{loss3}
\Delta \mathcal{L}_{b}(w) = \frac{1}{N - (b - 1)n_{c}} \sum \limits_{x_{i} \in \Delta \mathcal{X}_{b}} \ell(w, x_{i}).
\end{equation}
\normalsize
Note that we have the identity \small $\mathcal{L}(w) = \big((b - 1)n_{c}/N\big)\tilde{\mathcal{L}}_{b}(w) + \big((N - (b - 1)n_{c})/N \big)\Delta \mathcal{L}_{b}(w)$. 
\normalsize

First, we observe that given the previously transmitted data samples, the gradient $\nabla \ell(w_{b}^{j-1}, \xi_{b}^{j})$ is an unbiased estimate of the gradient $\nabla \tilde{\mathcal{L}}_{b}(w)$ of  the empirical loss limited to the samples available at the edge node at block $b$. In formulas, \small $\mathrm{E}_{\xi_{b}^{j} | \tilde{\mathcal{X}}_{b}}[\nabla \ell(w, \xi_{b}^{j})] = \nabla \tilde{\mathcal{L}}_{b}(w) $, \normalsize where \small $\mathrm{E}_{\xi_{b}^{j} | \tilde{\mathcal{X}}_{b}}[\ \cdot \ ]$ \normalsize is the conditional expectation given the previously transmitted samples. We finally make the following assumption (see, e.g., \cite{bottou}):

\begin{enumerate}[label=(\subscript{A}{\arabic*}), align=left, leftmargin=0.5\parindent]
\setcounter{enumi}{3}
    \item \label{ass4}  For any set $\tilde{\mathcal{X}}_{b}$ of samples available at the edge node, there exist scalars $M \geq 0$ and $M_{V} \geq 0$ such that
    \small
    \begin{align}
    \label{var}
    \mathrm{V}_{\xi_{b}^{j} | \tilde{\mathcal{X}}_{b}}[\nabla \ell(w, \xi_{b}^{j})] \leq M + M_{V}||\nabla \tilde{\mathcal{L}}_{b}(w)||_{2}^{2}
    \end{align}
    \normalsize
    where  \small$\mathrm{V}[\ \cdot \ ] = \mathrm{E}[||\cdot||^{2}] - ||\mathrm{E}[\ \cdot \ ]||^{2}$ \normalsize is the variance.
\end{enumerate}

\section{Convergence analysis}
\label{sec4}

In this section, we present our main result and its implications on the optimal choice of the number $n_{c}$ of transmitted samples per block. Henceforth, we use the notation \small $\mathrm{E}_{b}[\ \cdot \ ]$ \normalsize to indicate the conditional expectation  \small  $\mathrm{E}_{\xi_{b}^{1}, \dots \xi_{b}^{n_{p}}|\tilde{\mathcal{X}}_{b}}[\ \cdot \ ]$ \normalsize on the samples selected for the SGD updates in the $b$-th block given the set \small $\tilde{\mathcal{X}}_{b}$ \normalsize of samples available at the edge node at $b$. We similarly define \small $\mathrm{E}_{B_{l}}[\ \cdot \ ] = \mathrm{E}_{\xi_{B_{l}}^{1}, \dots, \xi_{B_{l}}^{n_{l}}}[\ \cdot \ ]$ \normalsize as the conditional expectation on the samples selected for the SGD updates in block $B_{l}$ (see Fig. \ref{fig 2}(b)). 
\vspace{10pt}
\begin{Theorem}
\label{th1}
Under assumptions \ref{ass1}-\ref{ass4}, assume that the SGD stepsize $\alpha$ satisfies
\small
\begin{align}
        0 < \alpha \leq \frac{2}{LM_{G}}
\end{align}
\normalsize
and define
\small
\begin{align}
        \gamma = \alpha\Big(1 - \frac{1}{2}\alpha L M_{G}\Big).
\end{align}
\normalsize
Then, for any sequence $\tilde{\mathcal{X}}_{1}, \dots, \tilde{\mathcal{X}}_{B}$ the expected optimality gap at time $T$ is upper bounded as

\small
\begin{align}
\label{res1}
 & \mathrm{E}_{B}[\mathcal{L}(w_{B}^{n_{p}}) - \mathcal{L}(w^{*})] & \nonumber \\
    &\leq \frac{\alpha^{2} L M}{2 \gamma c} \frac{(B - 1)}{B_{d}}  + \Big(1 - \frac{(B - 1)}{B_{d}}\Big) \mathrm{E}_{B}\Big[\Delta \mathcal{L}_{B}(w_{B}^{n_{p}}) - \Delta \mathcal{L}_{B}(w^{*})\Big]  + \frac{1}{B_{d}}\sum_{l=1}^{B - 1} (1 - \gamma c)^{l n_{p}}\mathrm{E}_{B - l}\Big[\mathcal{L}_{B - l}(w_{B - l}^{n_{p}}) - \mathcal{L}_{B - l}(w^{*}) - \frac{\alpha^{2} L M}{2 \gamma c}\Big]&
    \end{align}
\normalsize
if $T \leq B_{d}(n_{c} + n_{o})$; and by 
\small
\begin{align}
    \label{res2}
    & \mathrm{E}_{B_{l}}\Big[\mathcal{L}\Big(w_{B_{l}}^{n_{l}}\Big) - \mathcal{L}(w^{*})\Big] \leq \frac{\alpha^{2} L M}{2 \gamma c} + \frac{1}{B_{d}}(1 - \gamma c)^{n_{l}}\sum_{l= 0}^{B_{d} - 1} (1 - \gamma c)^{l n_{p}} \mathrm{E}_{B_{d} - l}\Big[\mathcal{L}_{B_{d} -l}(w_{B_{d} - l}^{n_{p}}) -  \mathcal{L}_{B_{d} -l}(w^{*}) - \frac{\alpha^{2} L M}{2 \gamma c}\Big] & 
\end{align}
\normalsize
if $T > B_{d}(n_{c} + n_{o})$. \\
\textit{Proof}: See Appendix A.
\end{Theorem}
\vspace{5pt}

The bound (\ref{res1})-(\ref{res2}) extends the classical analysis of the convergence of SGD for the case in which the entire dataset is available at the learner \cite[Theorem~4.6]{bottou} to the set up under study. The bound distinguishes the case in which the edge node has the entire data set by the last block, and the complementary case, as seen in Fig. \ref{fig 2}.

The first term in the bound  (\ref{res2}) represents an asymptotic bias that does not vanish with the number of SGD updates, even when all the data points are available at the edge node. It is due to the variance (\ref{var}) of the stochastic gradient. The bound (\ref{res1}) for smaller values of $T$ also comprises an additional bias term, that is the second term in (\ref{res2}), due to the lack of knowledge about samples not received at the edge node by the end of the training process. In contrast, the last term in bound (\ref{res1})-(\ref{res2}) accounts for the standard geometric decrease of the initial error in gradient-based learning algorithms. Here, the initial error for each block $b$ is given by $\mathrm{E}_{b}\big[\mathcal{L}(w_{b - 1}^{n_{p}}) - \mathcal{L}(w^{*})\big]$. Note that the additional factor with exponent $n_{l}$ in (\ref{res2}) accounts for the number of updates made after all the samples have been received at the edge node. 

The bound (\ref{res1})-(\ref{res2}) can be in principle optimized numerically in order to find an optimal value to the block size $n_{c}$. However, in practice, doing so would require fixing the choice of the sequence $\tilde{\mathcal{X}}_{1}, \dots, \tilde{\mathcal{X}}_{B}$, and running Monte Carlo experiments for every randomly selected sample of the sequence of SGD updates (\ref{SGD}), which is computationally intractable. Therefore, in the following, we derive a generally looser bound that can be directly evaluated numerically without running any Monte Carlo simulations. This bound will then be used in order to obtain an optimized value for $n_{c}$.
\begin{figure}
    \centering
    \includegraphics[scale=0.33]{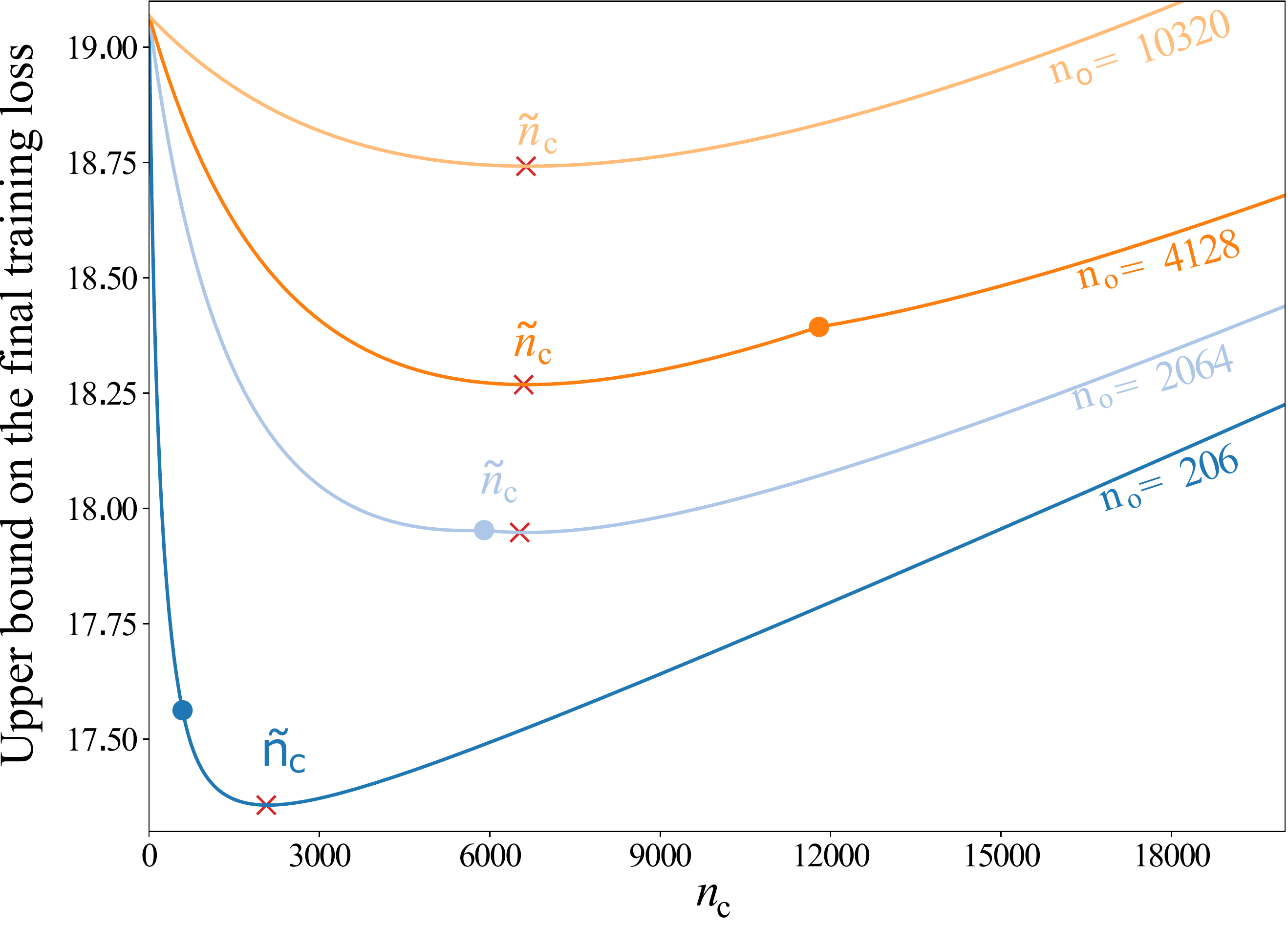}
    \caption{Upper bound (\ref{res3})-(\ref{res4}) versus block size $n_{c}$ for various values of the overhead $n_{o}$. The full dots represent values of $n_{c}$ at which we have $T = B_{d}(n_{c} + n_{o})$ (see Fig. \ref{fig 2}), crosses represent the optimized value $\tilde{n}_{c}$.}
    \label{simu 0}
\end{figure}
\vspace{10pt}
\begin{corollary}
\label{corr1}
Under the conditions of Theorem \ref{th1}, the expected optimality gap at time $T$ is upper bounded as
\small
\begin{align}
    \label{res3}
 & \mathrm{E}_{B}[\mathcal{L}(w_{B}^{n_{p}}) - \mathcal{L}(w^{*})] \leq \frac{\alpha^{2} L M}{2 \gamma c} \frac{(B - 1)}{B_{d}} + \Big(1 - \frac{(B - 1)}{B_{d}}\Big) \frac{LD^{2}}{2} + \frac{1}{B_{d}}\sum_{l=1}^{B - 1} (1 - \gamma c)^{l n_{p}}\Big[\frac{LD^{2}}{2} - \frac{\alpha^{2} L M}{2 \gamma c}\Big], &
    \end{align}
\nopagebreak
\normalsize
\nolinebreak
if $T \leq B_{d}(n_{c} + n_{o})$; and by
\end{corollary}
\small
\begin{align}
    \label{res4}
    & \mathrm{E}_{B_{l}}[\mathcal{L}(w_{B_{l}}^{n_{l}})  - \mathcal{L}(w^{*})] \leq \frac{\alpha^{2} L M}{2 \gamma c} + \frac{1}{B_{d}}(1 - \gamma c)^{n_{l}}\sum_{l= 0}^{B_{d} - 1} (1 - \gamma c)^{l n_{p}}\Big[\frac{LD^{2}}{2} - \frac{\alpha^{2} L M}{2 \gamma c}\Big] & 
\end{align}
\normalsize
if $T > B_{d}(n_{c} + n_{o})$. \\
\textit{Proof}: See Appendix B.
\vspace{5pt}

We plot bound (\ref{res3})-(\ref{res4}) in Fig. \ref{simu 0}. These results are obtained for $N = 18,576$, $T = 1.5 N$, $L = 1.908$, $c= 0.061$, $M = 1$, $M_{G}=1$, $\tau_{p}=1$, $\alpha = 0.0001$. We note that $L$  and $c$ represent respectively the smallest and largest eigenvalues of the data Gramian matrix for the example studied in Sec. \ref{simus}. For each value of $n_{o}$, we mark in the figure both the value of $n_{c}$ that minimizes the upper bound in Corollary \ref{corr1} and the value of $n_{c}$ at which we have the condition $T=B_{d}(n_{c} + n_{o})$. As seen in Fig. \ref{fig 2}, this is the minimum value of $n_{c}$ that allows the full transmission of the training set by the last training block.

A first observation is that the optimized value of $n_{c}$, henceforth referred to as $\tilde{n}_{c}$, is generally smaller than the number $N$ of training points in $\mathcal{X}$, suggesting the advantages of pipelining communication and computation. Furthermore, as the overhead $n_{o}$ increases, it becomes preferable, in terms of the bound (\ref{res3})-(\ref{res4}), to choose larger values $\tilde{n}_{c}$ for the block size $n_{c}$. This is because a larger value of $n_{o}$ needs to be amortized by transmitting more data in each block, lest the transmission time is dominated by overhead transmission. Finally, for smaller values of $n_{o}$, the minimum $\tilde{n}_{c}$ of the bound is obtained when the entire data set is eventually transferred to the edge node, i.e., $T > B_{d}(n_{c} + n_{o})$, while the opposite is true for larger value of $n_{o}$. Interestingly, this suggests that it may be advantageous in terms of final training loss, to forego the transmission of some training points in exchange for more time to carry out training on a fraction of the data set.

\section{Numerical experiments}
\label{simus}

\begin{figure}
    \centering
    \includegraphics[scale=0.33]{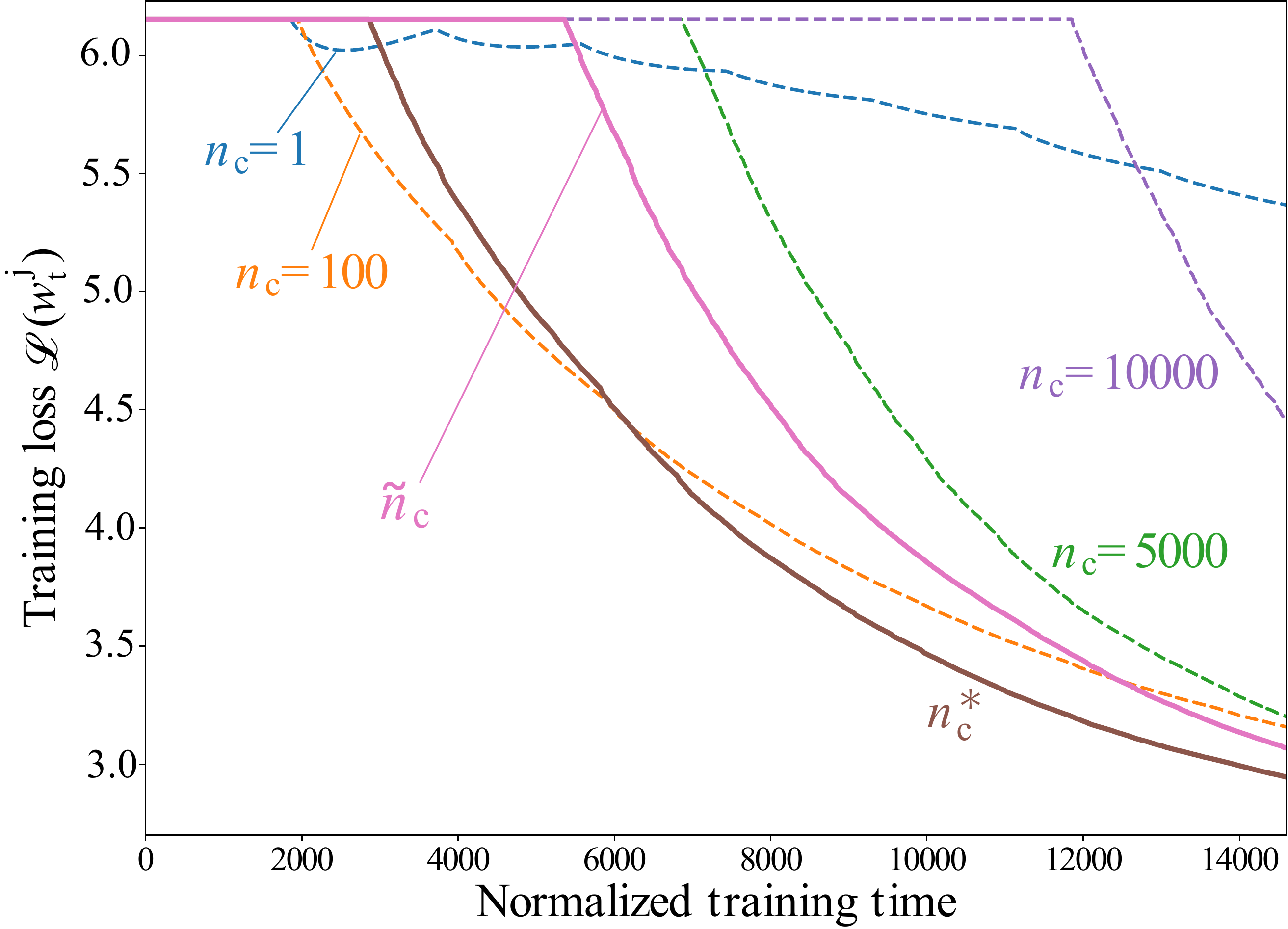}
    \caption{Training loss versus training time for different values of the block size $n_{c}$. Solid line: experimental and theoretical optima.}
    \label{simu 2}
\end{figure}

In this section, we validate the theoretical findings of the previous sections by means of a numerical example based on ridge regression on the California Housing dataset \cite{california}. The dataset contains 20640 covariate vectors $x_{n} \in \mathrm{R}^{8}$, each with a real label $y_{n}$. We randomly select $90\%$ of the samples to define the set $\mathcal{X}$ for training, i.e., we have $N = 18576$. As for Fig. \ref{simu 2}, we choose $\tau_{p} = 1$ and $\alpha=0.0001$. The parameter vector is initialized using i.i.d. zero-mean Gaussian entries with unitary power. The loss function is defined as \small
$\ell(w, x) = (w^{T}x - y)^{2} + \frac{\lambda}{N}||w||^{2}$
\normalsize where $w \in \mathrm{R}^{8}$ and the regularization coefficient is chosen as $\lambda = 0.05$. 

By computing the average final training loss for each value of $n_{c}$, we can experimentally determine the optimal value $n_{c}^{*}$ of the block size. 
We compare the performance using this experimental optimum with the performance obtained using the minimum $\tilde{n}_{c}$ of the bound (\ref{res3})-(\ref{res4}). 
To this end, in Fig. \ref{simu 2}, given a fixed overhead size $n_{o}$, we plot the average training loss $\mathcal{L}(w_{b}^{j})$ against the normalized training time $j$ for $n_{c}^{*}$ and for the value $\tilde{n}_{c}$ obtained from the bound (\ref{res3})-(\ref{res4}). As references, we also plot as dotted lines the losses obtained for selected values of $n_{c}$. The choice of the block size $n_{c}$ minimizing the average final loss is seen to be a trade-off between the rate of decrease of the loss and the final attained accuracy. In particular, decreasing $n_{c}$ allows the edge node to reduce the loss more quickly, albeit with noisier updates and at the cost of a potentially larger final training loss due to the transmitted packet being dominated by the overhead. Importantly, determining the optimum block size experimentally instead of using bound (\ref{res3})-(\ref{res4}) only provides a gain of $3.8\%$ in terms of the final training loss, at the cost of a computationally burdensome parameter optimization.
\section{Conclusions}
\label{sec6}
In this work, we considered an edge computing system in which an edge learner carries out training over a limited time period while receiving the training data from a device through a communication link. Considering a strategy that allows communication and computation to be pipelined, we have analysed the optimal communication block size as a function of the packet overhead. Among interesting directions for future work, we mention the inclusion of the effect of delays due to errors in the communication channel. In this case, the optimization problem could be generalized to account for the selection of the data rate. Other interesting extensions would be to consider online learning, where data sent in previous packets can be only partially stored at the server, and to investigate a scenario with multiple devices.

\clearpage
\appendices
\section{Proof of Theorem 1}
Using the same arguments as in the proof of  \cite[Theorem~4.6]{bottou}, we can directly obtain the following inequality for each block $b$:
\small
\begin{align}
\label{primarybottou}
    & \mathrm{E}_{b}[\tilde{\mathcal{L}}_{b}(w_{b}^{n_{p}}) - \tilde{\mathcal{L}}_{b}(w^{*})] \leq \frac{\alpha^{2} L M}{2 \gamma c}  +  (1 - \gamma c)^{n_{p}} \mathrm{E}_{b}\Big[\tilde{\mathcal{L}}_{b}(w_{b}^{0}) - \tilde{\mathcal{L}}_{b}(w^{*}) - \frac{\alpha^{2} L M}{2 \gamma c}\Big]. &
\end{align}
\normalsize
Note that we have $w_{b}^{0} = w_{b-1}^{n_{p}}$, since the initial parameter at block $b$ is the final parameter obtained at block $b-1$. By definition of the local empirical losses (\ref{loss1})-(\ref{loss2}), we have the equality
\small
\begin{align}
\label{surrloss}
\tilde{\mathcal{L}}_{b}(w_{b-1}^{n_{p}}) = \frac{b - 2}{b - 1}\tilde{\mathcal{L}}_{b-1}(w_{b-1}^{n_{p}}) + \frac{1}{b  - 1}\mathcal{L}_{b-1}(w_{b-1}^{n_{p}}).
\end{align}
\normalsize
Plugging (\ref{surrloss}) into (\ref{primarybottou}), we have
\small
\begin{align}
   \label{firststep}
   & \mathrm{E}_{b}[\tilde{\mathcal{L}}_{b}(w_{b}^{n_{p}}) - \tilde{\mathcal{L}}_{b}(w^{*})] & \nonumber\\
    & \leq \frac{\alpha^{2} L M}{2 \gamma c}  + (1 - \gamma c)^{n_{p}}\mathrm{E}_{b}\Big[\Big(
        \frac{b-2}{b-1}\Big)\Big(
            \tilde{\mathcal{L}}_{b-1}(w_{b-1}^{n_{p}})
             - \tilde{\mathcal{L}}_{b-1}(w^{*})\Big) + \frac{1}{b-1}\Big( \mathcal{L}_{b-1}(w_{b-1}^{n_{p}}) 
            - \mathcal{L}_{b-1}(w^{*}) \Big)
             - \frac{\alpha^{2} L M}{2 \gamma c}\Big]. &
 \end{align}
\normalsize
Iterating this substitution for all blocks $b-1, b-2, \dots, 2$, we obtain
\small
\begin{align}
    \label{basecase}
    & \mathrm{E}_{b}[\tilde{\mathcal{L}}_{b}(w_{b}^{n_{p}}) - \tilde{\mathcal{L}}_{b}(w^{*})] \leq \frac{\alpha^{2} L M}{2 \gamma c} + \sum_{l=1}^{b-1} (1 - \gamma c)^{l n_{p}}\frac{1}{b-1}\mathrm{E}_{b}\Big[
            \mathcal{L}_{b-l}(w_{b-l}^{n_{p}}) 
            - \mathcal{L}_{b-l}(w^{*})
            - \frac{\alpha^{2} L M}{2 \gamma c}\Big]. &
\end{align}
\normalsize
While inequality (\ref{basecase}) applies for any choice of $T$, we now specialize the result to the case where the allocated amount of time $T$ is not sufficient to transmit the whole dataset, i.e., $T \leq B_{d}(n_{c} + n_{o})$. (see Fig. \ref{fig 2}(a)).
According to (\ref{loss1})-(\ref{loss3}), for this case, we have the equality
\small
\begin{align}
\label{lossdecomp}
\mathcal{L}(w) = \frac{(b- 1)}{B_{d}}\tilde{\mathcal{L}}_{b}(w) + \frac{N - (b- 1)}{B_{d}} \Delta \mathcal{L}_{b}(w).   
\end{align}
\normalsize
Plugging (\ref{lossdecomp}) into (\ref{basecase}) for block $b=B$, we then obtain 
\small
\begin{align}
\label{result1th1}
 & \mathrm{E}_{B}[\mathcal{L}(w_{B}^{n_{p}}) - \mathcal{L}(w^{*})] & \nonumber \\
    &\leq \frac{\alpha^{2} L M}{2 \gamma c} \frac{(B - 1)}{B_{d}}  + \Big(1 - \frac{(B - 1)}{B_{d}} \Big) \mathrm{E}_{b}\Big[\Delta \mathcal{L}_{B}(w_{B}^{n_{p}}) - \Delta \mathcal{L}_{B}(w^{*})\Big] + \frac{1}{B_{d}}\sum_{l=1}^{B-1} (1 - \gamma c)^{l n_{p}}\mathrm{E}_{B}\Big[
            \mathcal{L}_{B - l}(w_{B - l}^{n_{p}}) 
            - \mathcal{L}_{B - l}(w^{*})
            - \frac{\alpha^{2} L M}{2 \gamma c}\Big], &
    \end{align} 
\normalsize
which is (\ref{res1}) in Theorem \ref{th1}.

Finally, we consider the case where there is sufficient time to transmit the whole dataset, i.e., $T > B_{d}(n_{c} + n_{o})$ (see Fig. \ref{fig 2}(b)).
According to (\ref{primarybottou}), we have
\small
\begin{align}
    \label{2ndcase}
 & \mathrm{E}_{B_{l}}[\mathcal{L}_{B_{l}}(w_{B_{l}}^{n_{l}}) - \mathcal{L}_{B_{l}}(w^{*})] & \nonumber \\
    &  \leq \frac{\alpha^{2} L M}{2 \gamma c}  + (1 - \gamma c)^{n_{l}}\mathrm{E}_{B_{l}}\Big[
            \mathcal{L}(w_{B_{l}}^{0}) 
            - \mathcal{L}(w^{*}) - \frac{\alpha^{2} L M}{2 \gamma c}\Big] & \nonumber \\
    & \stackrel{\text{(a)}}{\leq}\frac{\alpha^{2} L M}{2 \gamma c} + \frac{1}{B_{d}}(1 - \gamma c)^{n_{l}}\sum_{l= 0}^{B_{d} - 1} (1 - \gamma c)^{l n_{p}} \mathrm{E}_{B_{l}}\Big[\mathcal{L}_{B_{d} -l}(w_{B_{d} - l}^{n_{p}}) -  \mathcal{L}_{B_{d} -l}(w^{*}) - \frac{\alpha^{2} L M}{2 \gamma c}\Big], & 
\end{align}
\normalsize
where (a) arises from plugging (\ref{result1th1}) in  (\ref{2ndcase}) with $B = B_{d}$.
This is (\ref{res2}) in Theorem \ref{th1}, concluding the proof.

\section{Proof of Corollary 1}
Defining for all $t = 1,\dots, B_{d}$, the optimum solution \small $\Delta w_{b}^{*} = \argmin_{w} \Delta \mathcal{L}_{b}(w)$, \normalsize we can write
\small $\Delta \mathcal{L}_{b}(\Delta w_{b}^{*}) \leq \Delta \mathcal{L}_{b}(w^{*})$ \normalsize,
and hence also the inequality
\small
\begin{align}
&\Delta \mathcal{L}_{b}(w_{b}^{n_{p}}) - \Delta \mathcal{L}_{b}(w^{*}) \leq \Delta \mathcal{L}_{b}(w_{b}^{n_{p}}) - \Delta \mathcal{L}_{b}(\Delta w_{b}^{*}).&
\end{align}
\normalsize
Writing the Lipschitz continuity property of the gradients \ref{ass2} with \small$\nabla(\Delta \mathcal{L}_{b}(\Delta w_{b}^{*})) = 0$ \normalsize and \ref{ass1}, we have \small$\Delta \mathcal{L}_{b}(w_{b}^{n_{p}}) - \Delta \mathcal{L}_{b}(\Delta w_{b}^{*}) \leq \frac{LD^{2}}{2}$.
\normalsize
Using a similar argument, we can write \small$\mathcal{L}_{b}(w_{b}^{n_{p}}) -  \mathcal{L}_{b}( w_{b}^{*}) \leq \frac{LD^{2}}{2}$, \normalsize where \small $w_{b}^{*} = \argmin_{w} \mathcal{L}_{b}(w)$. \normalsize
\normalsize
Plugging this into (\ref{result1th1}), we obtain the inequality
\small
\begin{align}
 & \mathrm{E}_{B}[\mathcal{L}(w_{B}^{n_{p}}) - \mathcal{L}(w^{*})] \leq \frac{\alpha^{2} L M}{2 \gamma c} \frac{(B- 1)}{B_{d}}  + \Big(1 - \frac{(B - 1)}{B_{d}} \Big)\frac{LD^{2}}{2} + \frac{1}{B_{d}}\sum_{l=1}^{B-1} (1 - \gamma c)^{l n_{p}}\Big[
            \frac{LD^{2}}{2}
            - \frac{\alpha^{2} L M}{2 \gamma c}\Big], &
    \end{align} 
\normalsize
which is (\ref{res3}) in Corollary \ref{corr1}.
Following the same approach with (\ref{2ndcase}), we obtain
\small
\begin{align}
    & \mathrm{E}_{B_{l}}[\mathcal{L}(w_{B_{l}}^{n_{l}}) - \mathcal{L}(w^{*})] \leq \frac{\alpha^{2} L M}{2 \gamma c}  + \frac{1}{B_{d}}(1 - \gamma c)^{n_{l}}\sum_{l= 0}^{B_{d} - 1} (1 - \gamma c)^{l n_{p}} \Big[\frac{LD^{2}}{2} - \frac{\alpha^{2} L M}{2 \gamma c}\Big], & 
\end{align}
\normalsize
which is (\ref{res4}) in Corollary \ref{corr1}, completing the proof.

\end{document}